%% file: main.tex
\documentclass[a4paper]{article}
\usepackage{graphicx} % Required for inserting images
\usepackage[dvipsnames]{xcolor}
\usepackage{amsmath}
\usepackage{amssymb}
\usepackage[greek,english]{babel}
\usepackage{csquotes}
\usepackage{url}

\usepackage[margin=25mm]{geometry}
\usepackage{booktabs} 
\usepackage{csvsimple}

\title{Technical Report for the Forgotten-by-Design Project:\\Targeted Obfuscation for Machine Learning}

\author{Rickard Brännvall, Laurynas Adomaitis, Olof Görnerup and Anass Sedrati \\ \small Computer Science Department, Digital Systems, RISE Research Institutes of Sweden}
\date{December 2024}

\usepackage{xcolor}

\newcommand{\IN}{\mathrm{IN}}
\newcommand{\OUT}{\mathrm{OUT}}

\newcommand{\singlewidth}{0.8\textwidth}

\newif\ifmycondition
\myconditiontrue % Set the condition to true
\begin{document}

\maketitle

\begin{abstract}
\input{abstract}   
\end{abstract}

\tableofcontents

\input{intro}   
\input{leakpro}   
\input{body}

\clearpage
%\printbibliography
\bibliographystyle{unsrt}
\bibliography{main}

\clearpage
\input{appendix}   

\end{document}

%% file: abstract.tex
The right to privacy, enshrined in various human rights declarations, faces new challenges in the age of artificial intelligence (AI). This paper explores the concept of the “Right to be Forgotten” (RTBF) within AI systems, contrasting it with traditional data erasure methods. We introduce “Forgotten by Design,” a proactive approach to privacy preservation that integrates instance-specific obfuscation techniques during the AI model training process. Unlike machine unlearning, which modifies models post-training, our method prevents sensitive data from being embedded in the first place. Using the LIRA membership inference attack, we identify vulnerable data points and propose defenses that combine additive gradient noise and weighting schemes. Our experiments on the CIFAR-10 dataset demonstrate that our techniques reduce privacy risks by at least an order of magnitude while maintaining model accuracy (at 95\% significance). 
Additionally, we present visualization methods for the privacy-utility trade-off, providing a clear framework for balancing privacy risk and model accuracy.
This work contributes to the development of privacy-preserving AI systems that align with human cognitive processes of motivated forgetting, offering a robust framework for safeguarding sensitive information and ensuring compliance with privacy regulations.

%% file: intro.tex
\section{Introduction}
The right to privacy or private life is enshrined in the Universal Declaration of Human Rights (Article 12), the European Convention of Human Rights (Article 8), and the European Charter of Fundamental Rights (Article 7). However, the meaning of privacy protection changes with the technological landscape, and the enforcement of certain privacy principles varies. Artificial intelligence (AI) has reintroduced many themes initially addressed by the EU’s General Data Protection Regulation (GDPR). For example, the right to be forgotten has long been interpreted as the right to ask for the removal of personal data. 

However, AI models typically do not store data per se. They are trained on the data, and removing personal data from the original training dataset would not affect the model, which may still embed sensitive information. A famous instance of this dilemma is clearview.ai, which used images of people’s faces to train facial recognition models that remain operational despite the efforts of many individuals and civil society groups to have their faces removed from the model. The national regulators have fined the company in Europe, but the product remains available, and public versions of the same concept are now freely available.

To counteract the non-consensual use of personal data, activist groups and researchers have developed techniques that fall under the general term “obfuscation.” The FAWKES project provided a tool to combat clearview.ai specifically by poisoning their publicly scraped data. Brunton and Nissenbaum have popularized the idea of obfuscation in their seminal work “Obfuscation: A User's Guide for Privacy and Protest.” They argue for obfuscation as a way to take back control of one’s privacy.

The overarching goal of this paper is to implement the general idea of obfuscation in the design of privacy-preserving AI systems so that it is not a reactionary act of protest but rather a method to protect sensitive data points by design. 
We employ the concept of a privacy vulnerability score, which quantifies the susceptibility of individual data points to privacy attacks based on the differences in model behavior on in-training versus out-of-training data points.

With extensive testing, it is possible to validate that custom obfuscation added to a model prevents the inference of membership, thus satisfying most cases of the right to be forgotten.
To systematically evaluate the privacy risks associated with AI models, we employ Membership Inference Attacks (MIA) \cite{hu2022membership} in this work, specifically the LIRA attack \cite{carlini2022membership}. These attacks help identify whether specific data points were part of the training dataset, thereby assessing the model's vulnerability to privacy breaches. 
By integrating LIRA into our methodology, we aim to rigorously test and validate the effectiveness of our proposed obfuscation techniques in preserving privacy.

To carry out such work, we also need a machine learning task.
The CIFAR-10 dataset \cite{Krizhevsky2009Learning}, a widely used benchmark in machine learning research, consists of 60,000 32x32-pixel color images across 10 different classes. It is particularly useful for evaluating the performance and privacy risks of image classification models due to its diversity and complexity. ResNet-18 \cite{He2015Deep}, a deep residual network with 18 layers, is designed to address the vanishing gradient problem by using shortcut connections, making it a robust choice for image classification tasks.

\paragraph{Structure of the report. }
First, a general background and definitions are provided to set the context of this article. This includes an interpretation of the Right to Be Forgotten in AI environments and an explanation of machine unlearning (MU), which is one of the suggested ways to address it. A motivation is then shared as to why we decided to use another approach and its advantages compared to MU. This approach introduces elements (differential privacy, vulnerable data points, and membership inference attacks). The working method is then presented with its different steps, starting by setting metrics and identifying vulnerable data points. A numerical experiment is then conducted, leading to an analysis and discussion that concludes by demonstrating the effectiveness of our suggested techniques in reducing privacy risks while
maintaining model utility. 

\section{Background and Definitions}
\subsection{Understanding the Right to Be Forgotten (RTBF)}
\begin{quotation}
\textit{They are the souls who are destined for Reincarnation; and now at Lethe's stream they are drinking the waters that quench man's troubles, the deep draught of oblivion. They come in crowds to the river Lethe, so that you see, with memory washed out, they may revisit the earth above.} (Virgil, Aeneid 6. 705).
\end{quotation}
In these verses by Virgil, the father’s ghost explains to Aeneas one of the passages of souls from death to rebirth. According to the myth, every soul is made to drink from the river Lêthê (\textgreek{Ληθη}, forgetfulness) to forget their previous lives. Lêthê represents forgetting as an event, a sort of induced forgetting, that is brought upon people from the supernatural.

This mythical understanding is in contrast to the usual meaning of forgetting as a passive and systemic process of the mind, e.g., decay, that happens without intent or explicit planning. Forgetting can be described as a “brain-wide well-regulated decay process, occurring mostly during sleep, [that] systematically removes selected memories” \cite{hardt2013decay}. Other accounts of forgetting, like motivated forgetting, include an intentional element. It claims that “limiting awareness of unwanted memories makes us forget them” \cite{anderson2012towards}, and that is especially true for sensitive or traumatic events that people want to forget. However, motivated forgetting still accounts for forgetting by limiting one’s attention to the target memory. There does not seem to be a mechanism in human psychology that would be similar to the effect of the river Lêthê that makes one intentionally and actively forget\footnote{There is some promise from neurotechnologies for memory editing \cite{phelps2019memory}  but that these technologies are being developed only testifies to the fact that such editing does not occur naturally.}.

However, when human memory is metaphorically transferred to computing, it becomes like drinking from Lêthê. It is common to refer to a computer’s “memory” and to understand deleting data as the computer “forgetting” data. The origins of the metaphor of “computer memory” are found in von Neumann’s “First Draft of a Report on the EDVAC”\cite{von1993first}, which extensively makes use of the term “memory” in the definitions. It was part of a larger practice in early computing literature to use neurophysiological concepts to describe computer elements \cite{schmidt1998rechenmaschinenspeicher}. The notion of “computer memory” remained in use despite the technology undergoing significant changes \cite{magoun2015remembering}.

Such anthropomorphic metaphors are generally not problematic. In fact, they may be useful heuristics to help non-specialists understand technical concepts like computer storage without having to explain solid-state discs or similar technologies. In the previous decades, the careful and conscious curation of metaphors has helped “reduce the fear and ignorance that often dishearten first-time computer users” \cite{chisholm1986selecting}. A paradigm example of an endearing metaphor is the “computer mouse.”

However, there are risks associated with anthropomorphic metaphors. One such risk is that the overuse of anthropomorphism may lead to the disconnect between the intended metaphor and the original meaning. The disconnect is potentially harmful because the metaphor becomes uninformative and non-reflective, but it may also have effects on the original meaning of concepts, like human memory in “computer memory” or attention in the transformer revolution \cite{vaswani2017attention}. This phenomenon is called the bi-directionality of metaphors \cite{schoeneborn2013recontextualizing,wee2005constructing} or double mimetism \cite{grinbaum2023parole}. Metaphors create two-way dynamic relationships between source and target domains. The source domain can be shaped by the evolving target domain and vice versa. For example, using human cognition as a source domain helps illuminate computing and their "memory" or ability to "learn," but this also changes how we understand the human mind in view of computer function.

At least to some extent, this has happened in the discourse on computer memory and the “right to be forgotten”, which is enshrined in the article 17 of the GDPR. The article itself is called “Right to erasure (‘right to be forgotten’).” It is formulated as the data subject’s right “to obtain from the controller the erasure of personal data concerning him or her without undue delay, and the controller shall have an obligation to erase personal data without undue delay,” continuing to list grounds for such erasure. Here, rather than forgetting being a metaphor for erasure, erasure becomes the paradigm of forgetting. What the article requires is the technical process of erasing or deleting digital data, but that is also being taken as equivalent to the person’s right to be forgotten. This may seem innocuous \textit{prima facie}. However, it assumes a model of memory akin to the mythological drinking from Lêthê.

The erasure of digital data is an active and intentional process that requires finding and identifying the required data points and removing them from “memory” or storage. For example, in cybersecurity, one of the risk factors is residual evidence of sensitive information (e.g., cryptographic keys, personal identifiers), which often persists unexpectedly and requires actively finding and deleting these artifacts as a precaution \cite{shands2021intentional}. Likewise, executing the person’s right to erasure would require finding and wiping out the sensitive data, preferably using special methods, which do not rely on marking the space as free upon deletion but not actually changing the storage \cite{castiglione2011automatic}. However, this is not how forgetting happens in human cognition, neither under the decay model, which is passive, nor under-motivated forgetting, which relies on directing one’s attention away from the target memories.

This difference between erasure and forgetting becomes crucial in understanding the right to be forgotten in the context of artificial intelligence (AI), which employs a different mechanism of “memory” than traditional computing media. 
Not all AI systems learn from data; however, many methods that have contributed to the significant advancements in AI in recent years fall under the category of machine learning (ML), which focuses on developing algorithms that enable computers to learn from and make predictions based on data. Examples of ML models include neural networks, decision trees, and support vector machines. These models learn by adjusting their parameters through a process called training. During training, the model is exposed to a large dataset and iteratively updates its parameters to minimize the error between its predictions and the actual outcomes. Techniques such as gradient descent and backpropagation facilitate the model's gradual improvement by enabling it to learn patterns and relationships within the data. The learned parameters serve as a compressed representation of the training data, encapsulating the features necessary for the model to make accurate predictions. This mechanism of learning and storing information differs fundamentally from traditional data storage, where data is explicitly saved and can be directly accessed or deleted.

The evident problem with data deletion in AI models is that the learned parameters have already been extracted from the training data, and the training data is not directly stored within the model. Therefore, there seems to be nothing that could be deleted. Of course, the data subject’s request may trivially be addressed by deleting their part of the training data, but that does not change the model’s parameters unless the model is retrained. Retraining is often very resource-demanding, and it is not feasible to retrain a model just to remove a data point, so it would be highly impractical to understand the right to be forgotten as retraining.

Moreover, although the AI model does not directly store training data, it may recreate the original training data from its parameters. For example, it was possible to recreate sensitive data points, including accurate addresses and phone numbers in GPT2 \cite{carlini2021extracting}. It was also possible to extract a significant number of the original training texts from GPT3.5 turbo with a relatively low cost \cite{nasr2023scalable}. This reiterates that the right to be forgotten applies to AI models despite the fact that they do not directly store and delete in the way that traditional computer systems do.

\subsection{Privacy Risk Audits}

A privacy risk audit involves the systematic and reproducible testing of an AI system to ensure compliance with GDPR, the AI Act, and other related regulations. This process aims to verify that personally identifiable data is not leaked, whether through an API (black box scenario) or by a party with access to the learned model parameters (white box scenario). By conducting privacy risk audits, organizations can identify and mitigate potential privacy breaches, ensuring that their AI systems handle sensitive data responsibly and in accordance with regulatory requirements.

Membership inference attacks (MIAs) may play an important role in privacy risk audits. These attacks aim to determine whether a specific data point was part of the training dataset of a machine learning model. MIAs exploit the differences in how a model treats data it has seen during training versus data it has not seen. If an attacker can reliably infer the presence of specific data points in the training set, it indicates a privacy risk, as sensitive information about individuals could potentially be exposed.

In a recently published opinion \cite{EDPB2024}, the European Data Protection Board (EDPB) emphasizes the importance of assessing the residual likelihood of identification when evaluating the anonymity of AI models. According to the EDPB, AI models trained on personal data cannot always be considered anonymous, as information from the training dataset may still be extractable or otherwise obtained from the model. This highlights the need for thorough privacy risk assessments, including the use of MIAs, to ensure that AI models do not inadvertently expose personal data.

In its opinion, the EDPB outlines several key considerations for assessing the anonymity of AI models:

\begin{itemize}
    \item \textbf{Model Design}: Evaluating the approaches taken during the development phase to limit the collection of personal data, reduce identifiability, and prevent data extraction.
    \item \textbf{Data Preparation and Minimization}: Assessing the use of anonymization and pseudonymization techniques, as well as data minimization strategies.
    \item \textbf{Methodological Choices}: Ensuring the use of robust methods that reduce identifiability, such as differential privacy or other techniques for privacy risk mitigation.
    \item \textbf{Model Outputs}: Implementing measures to prevent the extraction of personal data from model outputs, such as output filters or post-training techniques like unlearning.
\end{itemize}

The EDPB also stresses the importance of documenting the measures taken to ensure anonymity and conducting regular assessments to verify their effectiveness. This documentation should include details on the technical and organizational measures implemented, the results of privacy risk assessments, and any mitigating measures taken to address identified risks.

\subsection{Research Question and Contribution}
If we cannot conceptualize the right to be forgotten through erasure, nor through retraining, nor can we be satisfied that AI models do not “remember” sensitive information by default, how should we understand this right in the context of AI? Contrary to recent approaches \cite{villaronga2018humans}, we suggest that forgetting in AI should be understood in a way similar to how humans forget rather than through erasure. It should be a motivated but systemic process that shifts the model’s parameters away from vulnerable data points, much like the paradigm of motivated forgetting in humans. Our proposed model consists of three major concepts based on human cognition - memory, forgetting, and asking. 

\begin{enumerate}
    \item Memory in AI is not based on direct or permanent storage, but on so-called parametric memory constituted by learned parameters and generalizations \cite{radford2019language}. Some AI models may be augmented with storage but that storage does not raise additional issues as compared to other standard computer media because they allow simple erasure. Thus, we define the memory of AI as contained in the learned parameters or generalizations rather than any additional storage. This definition also aligns with the phenomenology of memory, which acknowledges memory as a constructive activity which “acts to organize what might otherwise be a mere assemblage of contingently connected events” \cite{casey2000remembering}. The passivist model of memories as replicative repetitions of the past, which would approximate computer memory, has long been rejected in the cognitive and philosophical literature \cite{michaelian2018new,schacter2007cognitive}. Learned parameters in AI bear a much closer resemblance to the phenomenological concept of memory.
    \item The conception of memory based on parametric learning leads to a more nuanced version of forgetting that does not allow for simple erasure. Instead, it is modeled on the paradigm of motivated forgetting in human cognition, i.e., shifting the focus away or blurring our sensitive memories \cite{anderson2012towards} rather than direct deletion. We propose a method called Forgotten by Design that uses targeted obfuscation to achieve a level of non-inference, which we will equate to forgetting. In short, if there is no reliable (i.e., better than random) way of determining that a data point was part of the training data set, then we consider that the AI model has forgotten that data point during or after training.
    \item A key component of the Forgotten by Design model of AI memory is testing whether something has been forgotten. The test we adopt is called a membership inference attack (MIA) \cite{hu2022membership}, which aims to infer whether a data record was used to train a target model or not. This test replicates an indirect memory test in psychology \cite{mulligan1996divided} because the goal of an MIA is not to recover the specific data record itself but rather to determine whether a model exhibits a behavioral difference in processing a record that it has seen before. An indirect memory test in humans is also a test of the person’s capacity for implicit memory because it aims to measure the unintentional and unconscious influence of a prior experience (analogously, the training dataset in AI) on their behavior. 

\end{enumerate}

\begin{table}%[h]
\caption{Conceptual model for forgetting.}
\centering
\begin{tabular}{l l l} 
 \hline
 \textbf{Motivated forgetting in humans} & \textbf{Forgotten by Design model} & \textbf{Erasure model} \\ [0.5ex] 
 \hline%\hline
 Memory & Parameters & Training data/storage \\ 
 %\hline
 Forgetting & Obfuscation & Deletion \\
 %\hline
 Asking & Membership inference attack & Audit/crawling  \\[1ex] 
 \hline
\end{tabular}
\label{tab:conceptual_model}
\end{table}

\subsection{Difference with Machine Unlearning}
Machine unlearning \cite{xu2023machine} involves techniques for removing -- partly or in full -- sensitive information embedded in trained models without having to retrain the models from scratch. The goal, then, is to “untrain” a model in such a way that the resulting model behaves as a model that was trained from scratch without the unlearned examples. In the strongest case, the two models are statistically indistinguishable. However, in practice, this is often impossible to achieve, and less strict criteria are used, such as allowing that only subsets of activations (typically in the output layer) are statistically indistinguishable, at the cost of not guaranteeing that internal parameters are not unlearned. 

Machine unlearning is achieved by manipulating the training data, the model, or both, through obfuscation, pruning, or replacement of training examples or model parameters \cite{xu2023machine}. Model parameters may, for instance, be altered in such a way that they offset the effect of unlearned training examples \cite{graves2021amnesiac}. Another example is the approach to partition training data into subsets that are used to train submodels that have an aggregated output, e.g., through consensus \cite{bourtoule2021machine}. The advantage of this is that data pruning of sensitive data points will then only trigger re-training of associated submodels.

The key difference between machine unlearning and our approach is that the former attempts to modify model parameters or training data \emph{after} the sensitive information has already been incorporated into the model through training, whilst we use targeted obfuscation of sensitive information at the very outset, avoiding sensitive data to be embedded in the model in the first place. In this way, we achieve machine unlearning for any data point while training the original model, which makes further incremental unlearning unnecessary.

\subsection{Differential Privacy}

Differential Privacy (DP) is a technique that ensures the privacy of individual data points in a dataset by adding controlled noise to the data or computations, making it difficult to infer any single individual's information. 
For machine learning applications, it is implemented as a Differentially Private Stochastic Gradient Descent (DP-SGD), which is an algorithm designed to train machine learning models with strong privacy guarantees \cite{abadi2016dpsgd}. It modifies standard Stochastic Gradient Descent (SGD) by incorporating differential privacy techniques. During each iteration, DP-SGD computes the gradient of the loss function with respect to a mini-batch of training data. The algorithm clips the gradients to a fixed norm, limiting the influence of any single data point, and then adds Gaussian noise to the clipped gradients.  

The privacy guarantee is quantified by parameters $\epsilon$ and $\delta$, which control the trade-off between privacy and model utility. The final update step in DP-SGD can be expressed as:

\begin{equation}
\theta_{t+1} = \theta_t - \eta \left( \frac{1}{|B|} \sum_{i \in B} \left( \text{clip}(g_i, C) + \mathcal{N}(0, \sigma^2 C^2 I) \right) \right)
\label{eq:DPSGD}    
\end{equation}

where $\theta_t$ are the learned model parameters at iteration $t$, $\eta$ is the learning rate, $B$ is the mini-batch, $\text{clip}(g_i, C)$ is the gradient clipping function, and $\mathcal{N}(0, \sigma^2 C^2 I)$ represents the added Gaussian noise.

Rahimian et al. (2021) discuss various differential privacy (DP) defenses and their effectiveness against membership inference attacks, noting that while DP can robustly protect against strong adversaries, it often reduces model utility \cite{rahimian2021differential}. They emphasize that the noise added to ensure privacy can degrade model accuracy and performance, creating a challenging trade-off between privacy and utility. Their study underscores the importance of carefully tuning DP parameters to balance these competing objectives, as overly aggressive privacy settings can significantly impair model performance, while insufficient privacy measures may leave models vulnerable to attacks.

\subsection{Vulnerable data points}

In the context of membership inference attacks (MIAs), vulnerable data points are those that can be accurately identified as being part of a machine learning model's training dataset. MIAs exploit the differences in how a model treats data it has seen during training versus data it has not. If an attacker can determine whether a specific data point was used in training, they can infer sensitive information about that data point.
Vulnerable data points are particularly susceptible because they exhibit characteristics that make them easier to distinguish. For example, these points might be outliers or have unique features that the model memorizes more distinctly, which makes it easier for an attacker to infer their presence in the training set \cite{conti2022vulnerability}.

An effective method for identifying vulnerable data points is the LIRA membership inference attack \cite{carlini2022membership}. LIRA employs multiple shadow models trained on subsets of the dataset to estimate the likelihood that a specific data point was part of the training set. This method models the losses of these shadow models using Gaussian distributions, providing a robust estimation of privacy risks.

Carlini et al. introduced the concept of the Privacy Onion \cite{carlini2022privacyonion} to describe how removing the most vulnerable data points to privacy attacks exposes a new layer of previously safe data to the same risks. 

\subsection{Defenses against Privacy Attacks}

Earlier empirical defenses for machine learning privacy often forgo the provable guarantees of differential privacy (DP) to attain greater utility while resisting realistic adversaries. 
However, these defenses can exhibit significant weaknesses, as highlighted by \cite{misleading_defences}. The paper argues that previous evaluations often do not adequately capture the privacy leakage of the most vulnerable samples, employ weak attacks, and neglect comparisons with practical differential privacy (DP) baselines.
As a result, they can underestimate privacy leakage by orders of magnitude. 
Stronger evaluations reveal that none of these empirical defenses are competitive with a well-tuned, high-utility DP-SGD baseline, even when the formal DP guarantees are relatively weak.

The importance of using appropriate metrics in privacy evaluations cannot be overstated. Carlini et al. critique the traditional evaluation methodology \cite{carlini2022membership}, noting that it does not reflect an attacker’s ability to confidently breach the privacy of any individual sample. They propose measuring the true positive rate (TPR) at a low false positive rate (FPR) to better capture the attacker's precision. This approach has been adopted by many recent works, but it still aggregates TPR and FPR over a data population, failing to address the privacy of the most vulnerable samples.

Therefore, it is argued \cite{misleading_defences} that valuations should focus on vulnerable data points to provide a more accurate assessment of privacy risks. Differential privacy (DP) offers a robust framework by bounding the TPR-to-FPR ratio of any membership inference attack, providing worst-case guarantees for every dataset and target sample. This connection can be leveraged to use membership inference attacks to lower-bound the DP guarantees of an algorithm.
Thus, differential privacy, when calibrated to achieve a practical utility-privacy trade-off, emerges as a reasonable defense.

\subsection{Scope and Limitations}
The scope of the current paper is to explore the concept of the "right to be forgotten" in relation to an AI system. However, we would like to acknowledge a number of limitations that are related to our work that will be important to take into consideration for future work.
\begin{itemize}
    \item This work is experimented with a single dataset (CIFAR-10). There will be a need for further use cases to scale and expand the results.
    \item The current definition of the right to be forgotten in the GDPR\footnote{https://gdpr.eu/right-to-be-forgotten/} is limited to erasing data, which is something our work is contesting. It is possible that changes in future legislation will go against our suggested definitions that were made to address the current gap in RTBF for AI.
   \item The evaluation of this work is conducted using Membership Inference Attacks, specifically LIRA. Other attacks could be used, but they are beyond the current scope.
   %\item\colorbox{Yellow}{OTHER LIMITATIONS? (question for all :)}
\end{itemize}
A more detailed discussion about the limitations of this work is presented in Section \ref{sec:discussions} (Discussions). 

%% file: leakpro.tex
\subsection{Relation to the LeakPro Project}

The LeakPro project\footnote{https://www.vinnova.se/en/p/leakpro-leakage-profiling-and-risk-oversight-for-machine-learning-models/}, funded by Vinnova, aims to create a platform for evaluating the risk of information leakage in machine learning applications. This issue is recognized as a significant threat by regulators worldwide, including the Swedish Authority for Privacy Protection (IMY), as highlighted in their regulatory sandbox study on federated learning. This initiative is a collaborative effort involving industry and public sector partners such as AstraZeneca, Sahlgrenska, and Region Halland, working together with AI Sweden and RISE to ensure systematic and reproducible testing of AI systems. The platform will support various attack scenarios, including black-box, white-box, federated learning, and synthetic data, aligning with the requirements of the upcoming AI Act. By developing LeakPro as an open-source tool \cite{LeakPro2025}, the project seeks to provide scalable and relevant solutions for assessing and mitigating data leakage risks in real-world settings, thus helping Swedish industry and public sector to proactively design AI systems with data protection and prepare for future regulatory reporting requirements.

In the context of this report, work from LeakPro will be used to identify vulnerable data points in a benchmark dataset.
Subsequently, LeakPro is used to evaluate the efficacy of the proposed mitigation.

%% file: body.tex
\section{Methods}
The method applied in this study involves several key steps. Initially, a Membership Inference Attack (MIA) is used to identify vulnerable data points within a benchmark dataset. Following this, a privacy risk mitigation method is developed, which incorporates adding noise to the gradient sum after assigning lower weights to vulnerable data points identified by their privacy vulnerability t-score. Numerical experiments are then conducted on MIA benchmarks, specifically utilizing the LIRA method. The strength of the privacy protection provided by these mitigations is subsequently evaluated.

\paragraph{The Anatomy of Privacy Risk Audit.}
Conducting a privacy risk audit involves a systematic approach to evaluate and mitigate privacy risks associated with AI models. 
It is our intention to simulate some of the steps that would be conducted in a privacy risk audit.
The following steps outline the process that we used for our purposes:
\begin{enumerate}
    \item \textbf{Establish a Baseline}: Assess the privacy risk of the model without any privacy risk mitigation using an audit dataset equivalent to the training data.
    
    \item \textbf{Implement Risk Mitigation}: Apply privacy risk mitigation techniques that we propose in this work, namely a combination of additive noise and a weighting scheme. Retrain the model with these mitigations in place to ensure that the model incorporates the privacy-preserving measures.
    
    \item \textbf{Test the Protected Model}: Use the same privacy risk assessment methods employed in the baseline assessment to evaluate the protected model (to ensure consistency in the evaluation process).
    
    \item \textbf{Analyze the Results}: Compare the results of the protected model with the baseline assessment. Analyze the effectiveness of the risk mitigants in reducing privacy risks. Demonstrate the efficacy of the implemented mitigations by highlighting improvements in privacy risk metrics.
    
    \item \textbf{Document the Audit}: Thoroughly document all steps of the audit process, including the methodologies used, the implementation of risk mitigations, the results of the assessments, and a thorough discussion of the robustness and limitations of the methodology.
\end{enumerate}
We note that this is a limited version of what would be required in a real-world setting.

\subsection{The LIRA Membership Inference Attack}

The LIRA membership inference attack \cite{carlini2022membership} is a technique used to assess the privacy risks of machine learning models. It involves training multiple shadow models on subsets of the dataset and evaluating their performance on specific data points to infer whether those points were part of the training set. This method leverages differences in model behavior on in-training versus out-of-training data points, providing a detailed analysis of privacy vulnerabilities.

LIRA was chosen for privacy auditing due to its robustness and effectiveness in identifying vulnerable data points. LIRA's approach of modeling losses with Gaussian distributions and using likelihood ratio tests provides a precise method for detecting privacy risks. Its ability to handle large datasets and empirical validation in various studies make it suitable for our privacy auditing needs.

In LIRA, multiple shadow models are trained on subsets of the underlying dataset. 
Each model $i$ when evaluated on a data point $j$ produce an observations $z_{i,j}$, 
for example, 
the cross-entropy loss nor the logits.
The LIRA-attack models these losses by a Gaussian distribution, which is specific to whether the point was part of the training set, IN, or not, OUT. 
With the Gaussian approximation, only the mean and standard deviation need to be estimated for each.
However, since the losses are not well approximated by a Gaussian distribution, LIRA uses logit scaling to ensure support in $(-\infty, \infty)$:
$$
\varphi_{j} = 
\phi(f; x_j,y_j) = 
\log\left( \frac{p_{j}}{1-p_{j}} \right)%, \quad p_{j} = \exp(-z(f; x_j,y_j)).
$$
where $p_j$ is the predicted probability for the true label for data point $j$. There are empirical evidence in~\cite[Fig.~4]{carlini2022membership} that the quantity described by $\varphi_j$ is well approximated by a normal distribution. 

The attack then uses the likelihood ratio test as decision criteria:
\begin{equation}
\Lambda (f; x_j,y_j) 
= \frac{\mathcal{N}(\phi(f; x_j, y_j); m^{\mathrm{IN}}_{n,j}, V^{\mathrm{IN}}_{n,j})}{\mathcal{N}(\phi(f; x_j, y_j); m^{\mathrm{OUT}}_{n,j}, V^{\mathrm{OUT}}_{n,j})
}    
\label{eq:LIRAobservations}
\end{equation}
where we note that $n$ denotes the total number of shadow models trained.
We sometimes write this as the log-likelihood ratio instead
\begin{align}
\lambda_{j} &= \log \Lambda(f; x_j,y_j) \\
&= \log \mathcal{N}(\varphi_{j}; m^{\mathrm{IN}}_{n,j}, V^{\mathrm{IN}}_{n,j}) 
- \log\mathcal{N}(\varphi_{j}; m^{\mathrm{OUT}}_{n,j}, V^{\mathrm{OUT}}_{n,j}) 
\label{eq:loglikelihoodlira}
\end{align}

\iffalse
\paragraph{Variance estimation.}
The per data point sample estimate for variance can be relatively uncertain if few shadow models are used. For robustness, a fixed variance can be used that is shared by all data points (but different for IN and OUT models). As a rule of thumb, if $n<60$, we take the global variance estimate, 
%
$$
V^{\IN} = \frac{1}{|I^{\IN}| - 1} \sum_{j \in I^{\IN}} (z_j - \bar{z})^2
$$
%
where $I^{\mathrm{IN}}$ is the set of all data points that were in the training set for shadow model $i$, 
and where we have a corresponding formula for the OUT case. 
If we have $n>60$, we instead take the variance estimated only on observations for each specific data point. 
As there may be variation in the IN and OUT count for individual data points, one can also consider this rule applied on a datapoint-by-dataapoint basis. 
\fi

\paragraph{Example attack.}
To give a concrete example, we briefly discuss here a LIRA-based attack that targets a ResNet-18 model trained on the CIFAR-10 dataset, which exploits the model's predictions to infer sensitive information about the training data. The CIFAR-10 dataset consists of 60,000 32x32 color images across 10 classes, commonly used for image classification tasks \cite{Krizhevsky2009Learning}. 
ResNet-18 is a deep residual network with 18 layers, designed to address the vanishing gradient problem by using shortcut connections \cite{He2015Deep}. The attack is carried out by training a number, $n$, of shadow models that are each trained on subsets of the CIFAR-10 dataset. 
These shadow models are used to generate observations $\varphi_{i,j}$ for each data point $j$ using the logit scaling of equation (\ref{eq:LIRAobservations}). The LIRA attack models the observations $\varphi_{i,j}$ using Gaussian distributions, distinguishing between whether the data point was part of the training set (IN) or not (OUT) for each shadow model $i$. The attack then employs a likelihood ratio test to decide if a data point was part of the training set, using the log-likelihood ratio for more robust decision-making.

\subsection{Privacy risk metrics}

\begin{figure*}%[H]
\centering
\includegraphics[width=\singlewidth]{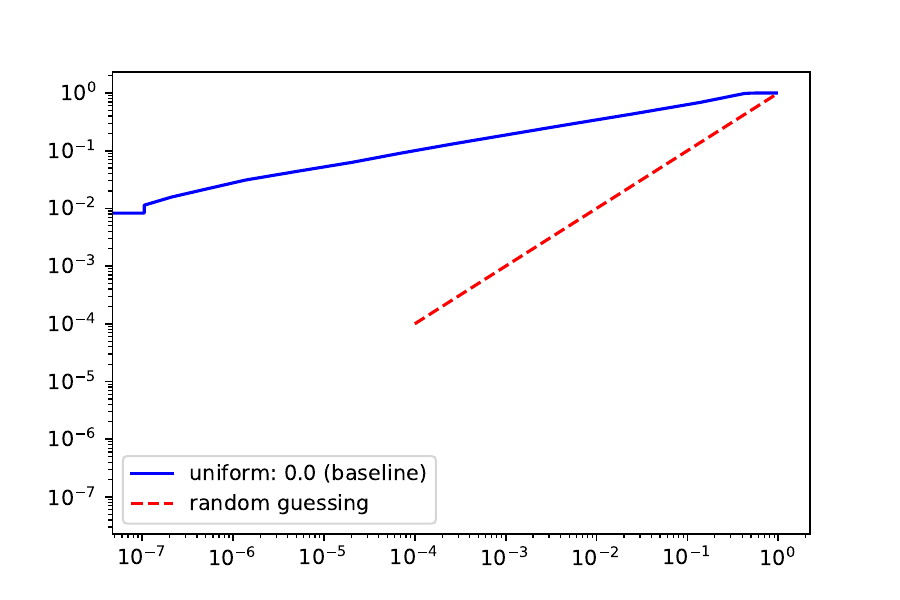}
\caption{ROC Curve, for example, LIRA Membership Inference Attack on ResNet-18 Model Trained on CIFAR-10 Dataset. The curve illustrates the True Positive Rate (TPR) against the False Positive Rate (FPR) for various threshold settings, demonstrating the model's ability to distinguish between members and non-members of the training dataset. The attack uses logit scaling and Gaussian distributions to model the observations, employing a likelihood ratio test for decision-making. The performance of the LIRA attack is compared to a random attack, showing a significant improvement in identifying training data points over random guessing}
\label{fig:roc_baseline}
\end{figure*}

The effectiveness of a membership inference attack (MIA) is commonly evaluated in terms of the ROC curve, AUC, and TPR at low FPR metrics:

\begin{description}
    \item[ROC Curve:] The Receiver Operating Characteristic (ROC) curve plots the True Positive Rate (TPR) against the False Positive Rate (FPR) at various threshold settings. It provides a visual representation of a model's ability to distinguish between members and non-members of the training dataset. Figure \ref{fig:roc_baseline} illustrates the ROC curve for the example attack against a ResNet-18 model trained on CIFAR-10 that we discussed in the previous section. 
    \item[AUC (Area Under the Curve):] The AUC is a single scalar value that summarizes the overall performance of the ROC curve. It ranges from 0 to 1, with a value closer to 1 indicating better performance. A higher AUC means the model has a better capability to differentiate between members and non-members. Although AUC is often used to express the privacy risk, it is important to note that it is an average metric and may not fully capture the nuances of privacy vulnerabilities.
    \item[TPR at a low FPR:] Evaluating TPR at a low FPR is particularly important for MIAs. This metric focuses on the attack's ability to correctly identify members while keeping false positives to a minimum. An attack that can reliably identify a few data points with high confidence (high TPR) at a low FPR is considered stronger than one that identifies many data points but with less reliability.
    \item[tau:] For this work we introduce tau, which is simply the logarithm of the ratio of TPR to FPR (at a specific FPR). It can interpreted as a lower bound on the $\varepsilon$ metric used for Differential Privacy.     
\end{description}

In the LIRA attack, these metrics are estimated by thresholding the likelihood ratio. The likelihood ratio is calculated for each data point, representing the probability of the data point being a member versus a non-member. By adjusting the threshold, one can control the trade-off between TPR and FPR. The ROC curve is generated by plotting TPR against FPR for different thresholds, and the AUC is computed as the area under this curve. To focus on the attack's strength, TPR at a low FPR is specifically examined, for example, TPR and 0.1\% FPR, as it highlights the attack's precision in identifying true members without falsely labeling non-members.

In summary, privacy should not be based on average metrics. Instead, MIAs should be evaluated based on their TPR at a low FPR. The rationale is that an attack that can reliably identify a few data points is stronger than an attack that unreliably identifies many~\cite{carlini2022membership}.

\paragraph{Attacks on shadow models.}
In addition to applying the LIRA attack only to the target model $f$, we can also apply it to each of the shadow models, $f_i$, for $i = 1, 2, \ldots, n$. 
$$
\varphi_{i,j} =  \phi(f_i; x_j,y_j) = \log\left( \frac{p_{i,j}}{1-p_{i,j}} \right), \quad p_{i,j} = \exp(-\ell(f_i; (x_j,y_j)))
$$
It is then appropriate to exclude observations from that model when estimating the loss mean and variance. Hence, we write
$$
\Lambda (f_i; x_j,y_j) = \frac{\mathcal{N}(\phi(f_i; x_j, y_j); m^{\mathrm{IN}}_{\text{-}i,j}, V^{\mathrm{IN}}_{\text{-}i,j})}{\mathcal{N}(\phi(f_i; x_j, y_j); m^{\mathrm{OUT}}_{\text{-}i,j}, V^{\mathrm{OUT}}_{\text{-}i,j})
}.
$$
where $\text{-}i$ is used to indicate parameters that are estimated on all shadow models except the $i$-th model.
Written as the log-likelihood ratio
\begin{align}
\lambda_{i,j} &= \log \Lambda(f_i; x_j,y_j) \\
&= \log \mathcal{N}(\varphi_{i,j}; m^{\mathrm{IN}}_{\text{-}i,j}, V^{\mathrm{IN}}_{\text{-}i,j}) 
- \log\mathcal{N}(\varphi_{i,j}; m^{\mathrm{OUT}}_{\text{-}i,j}, V^{\mathrm{OUT}}_{\text{-}i,j})
\end{align}
for which we now have a value for each data point for each of the $n$ shadow models. We can use this, for example, to construct $n$ ROC curves and estimate confidence bands on the corresponding privacy risk metrics.

\subsection{Identifying vulnerable data points}
The log-likelihood ratio $\lambda_j$ defined in the section above is used to decide if a data point $(x_j,y_j)$ was part of the training data set for a target model, given the mean and variance estimates we obtained from the shadow models. We expect that the estimated IN and OUT distributions for a data point will stabilize as we take more and more shadow models. 

We could consider using the log-likelihood ratio $\lambda_j$ for data point $j$ calculated by equation (\ref{eq:loglikelihoodlira}) as an estimate for the privacy attack vulnerability for the data point. 
However, as the estimated $\lambda_j$ depends on the target model, we could expect a considerably different value if we retrain the target model with a different random seed. 
To reduce the dependence on a single target model and its associated variability, we could instead use some measure of the average vulnerability for a single data point. For this purpose, we define the privacy vulnerability t-score  
\begin{equation}
t_j = \frac{m^{\mathrm{IN}}_j-m^{\mathrm{OUT}}_j}{\sqrt{V^{\mathrm{IN}}_j+V^{\mathrm{OUT}}_j}}
\label{eq:tscore}
\end{equation}
which only uses the means and variances estimated from the shadow models.

The privacy vulnerability t-score is related to the privacy risk defined in \cite{carlini2022membership}, which measures the vulnerability of data points to membership inference attacks. It is almost identical, except that our formulation expresses the denominator in terms of the variances in place of the sum of standard deviations used in the original reference. Our definition is altered to align with the standard formulation for the t-test for means of two normally distributed variables. 

Moreover, Choi et al. (2023) define the memorization score as a measure of how much a model's prediction for a data point improves when that point is included in the training set compared to when it is not. This score is based on the concept of memorization, where data points that are memorized by the model are more susceptible to such attacks. They demonstrate that by strategically selecting samples with high memorization scores, adversaries can maximize their attack success while minimizing the number of shadow models required \cite{Choi2023WhyTM}.

Alternatively, we could use a hold-one-out procedure to estimate the average value of AUC previously defined over all shadow models. The same averaging procedure can also be applied to tail risk metrics like TPR at low FPR, but these may have large variability unless we use many shadow models. For this work, however, we only employ the privacy vulnerability t-score to rank data points.

We employ the LeakPro software platform \cite{LeakPro2025} developed for quantitative privacy risk assessments machine learning models in our work because it integrates the LIRA membership inference attack. LeakPro implements the training of shadow models and the extraction of the required metrics, such as log-likelihood ratios and scaled logits. We can thus leverage the platform to identify vulnerable data points and evaluate the efficacy of the proposed privacy risk mitigation strategies.

\subsection{Defending vulnerable data points}
\label{sec:method_defending}

The approach explored in this paper uses an alternative version of the formula for the DG-SGD parameter update of equation (\ref{eq:DPSGD}) modified to make the obfuscation specific to the individual data point. We call this instance-specific gradient obfuscation, which we define as 
\begin{equation}
\theta_{t+1} = \theta_t - \eta \left( \frac{1}{|B|} \sum_{j \in B} \left( w_j g_j + u_j \right) \right)
\label{eq:DPSGD2}    
\end{equation}
with added noise $u_j \sim \mathcal{N}(0, \sigma^2_j I)$ that is allowed to depend on the individual data point, and $w_j$ is the data point specific weight. 

The effect of instance-specific gradient obfuscation is obtained by altering either
$\sigma_j$, such that a larger magnitude noise is applied for a vulnerable data point or
$w_j$, such that a vulnerable data point is down-weighted in the sum,  
either way reducing its relative contribution to the gradient update. 

We note that by not using gradient clipping, we forgo the formal guarantees of DP-SGD but argue that this may be reasonable when operating at noise levels that do not provide strong guarantees anyway. Gradient clipping can be computationally expensive and slow down training under DP-SGD, while our approach can be efficiently implemented simply as a weighted loss function.

\paragraph{Instance specific noise.}
One alternative approach could take $\sigma_j=(t_j)$ such that a point with a higher vulnerability score $t_j$ attracts an obfuscating noise of a larger magnitude
\begin{equation}
t_i \geq t_j \implies \sigma(t_i) \geq \sigma(t_j)
\label{eq:instance_noise}    
\end{equation}

This is akin to instance-specific differential privacy introduced by Nissim et al. (although they do not specifically discuss it in the context of machine learning or gradient descent) \cite{nissim2007smooth}.  

\paragraph{Instance specific weights.}
For this work, we instead consider the alternative method of keeping the noise constant,
$\sigma_j=\sigma^*$, 
while we let the weights depend on the vulnerability score
$w_j = w\left(t_j\right)$
through a non-increasing function 
\begin{equation}
%\forall t_i, t_j \in \mathbb{R}, \quad 
t_i \geq t_j \implies w(t_i) \leq w(t_j)
\label{eq:instance_weights}    
\end{equation}
There are, of course, many different possible functional relationships to consider -- linear, exponential, inverse, power laws, etc. For the experiments, we work with a clipped exponential
\begin{equation}
w^{\mathrm{clip}}(t) = \mathrm{clip}(
    e^{-\alpha t + \beta};
    w^{\mathrm{lower}}, w^{\mathrm{upper}}
)    
\end{equation}
where we furthermore require the weights to fall within $[0,1]$, which simplifies the weighting function to 
\begin{equation}
w^{\mathrm{exp}}(t) = e^{-\max(0,\alpha t - \beta)}
\label{eq:exp_weights}    
\end{equation}
for parameters $\alpha>0$ and $\beta\geq0$ that control the steepness and clipping point of the weight scaling.

\section{Numerical Experiments}

We experiment with a deep neural network model based on the ResNet architecture applied to the image classification task for the CIFAR-10 dataset. 
We use the model out-of-the-box as downloaded through the PyTorch library \cite{NEURIPS2019_9015}, but without pre-trained weights. 
We implement the instance-specific weighting of gradients as described in Section \ref{sec:method_defending}, equations (\ref{eq:DPSGD2}) and (\ref{eq:instance_weights}) and use the exponential weighting method of equation (\ref{eq:exp_weights}).
The experiments repeat the training of the model under different choices of the parameters $\sigma$, $\alpha$, and $\beta$. For clarity, we repeat the interpretation of these parameters in Table \ref{tab:control_parameters}.

\begin{table}%[h]
\caption{To reduce the threat from Membership Inference Attacks (MIAs), we propose a mitigation method that relies on noise injection and re-weighting of the data points based on their privacy risk vulnerability. The method relies on the three parameters $\sigma$, $\alpha$, and $\beta$ described in this table.}
\centering
\begin{tabular}{cl}
\hline
Parameter & Description \\
\hline
$\sigma$ & magnitude of the Gaussian noise injected into the gradients \\
$\alpha$ & controls the steepness by which the weights decrease with increasing privacy risk \\
$\beta$  & controls the cut-off point below which weights are equal to one \\
\hline
\end{tabular}
\label{tab:control_parameters}
\end{table}

Different parameter settings give us the opportunity to explore scenarios where the noise addition and weighting are altered. Mainly, we will be discussing the following alternatives 
\begin{description}
    \item[\textbf{baseline}:] training without any modifications to the gradient update, $\sigma=\alpha=\beta=0$
    \item[\textbf{uniform}:]  noise with $\sigma$ is added, but there is no weighting, $\sigma>0$, $\alpha=\beta=0$ 
    \item[\textbf{weighted}:] both noise addition and weighting are applied, $\sigma>0$, $\alpha>0$, $\beta\geq0$    
\end{description}

\paragraph{Main experiment set-up.}
The LIRA membership inference attack is carried out against the ResNet-18 model trained on the CIFAR-10 task, exploring different parameter choices for $\sigma$, $\alpha$, and $\beta$. The original train-test split of the dataset is maintained, with a test set of 10,000 images set aside solely for evaluating test accuracy. The remaining 50,000 images constitute the audit dataset. For each shadow model, the audit dataset is split into an IN partition, used for training, and an OUT partition, used only for evaluating the LIRA attack.

Initially, the baseline setting (no noise, no weights) is attacked to establish a baseline for the privacy metrics and to estimate the privacy vulnerability t-score for each image in the audit dataset. Up to 900 shadow models are used to ensure sufficient statistical validity for the subsequent robustness analysis. 

The experiments are then repeated for the uniform and weighted alternative scenarios, using more than 300 shadow models for each parameter setting. The weighted scenarios use the exponential weighting scheme defined in equation (\ref{eq:exp_weights}) based on the privacy vulnerability t-score calculated as in equation (\ref{eq:tscore}), that is, weights falling off exponentially fast with the (non-negative) distance between the estimated center of the IN and OUT distributions of the LIRA attack model.  

Finally, for each parameter setting, we evaluate the privacy metrics as well as the test set accuracy and store the results for analysis.
Figure \ref{fig:comparison} shows the resulting ROC curves, while Table \ref{table:results_summary} shows the AUC and $\tau$ (log TPR to FPR ratio) for a few selected parameter choices.

\subsection{Results}

\begin{figure*}%[H]
\centering
\includegraphics[width=\singlewidth]{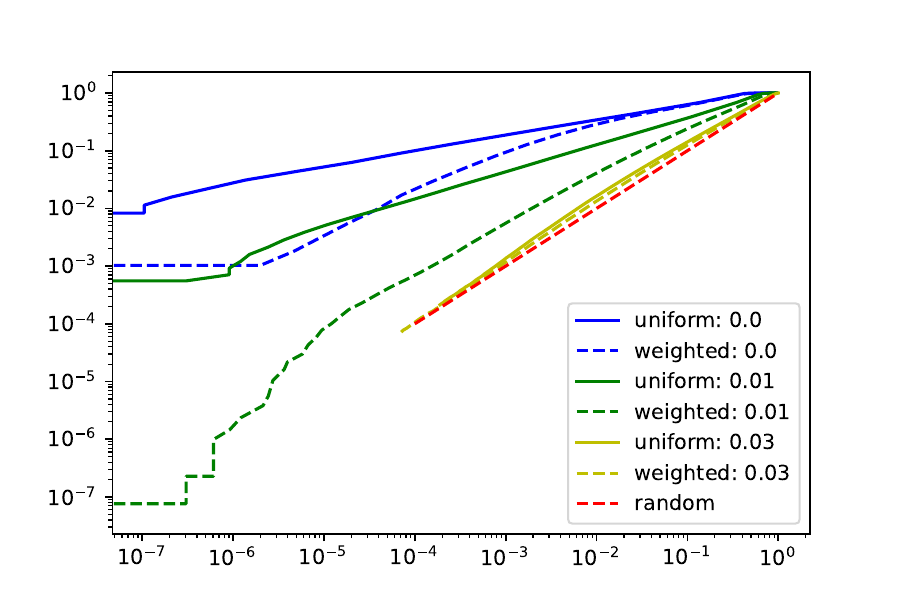}
\caption{ROC Curves for LIRA Membership Inference Attack on ResNet-18 Model Trained on CIFAR-10 Dataset comparing the performance of the attack under different scenarios: baseline ("uniform: 0.0") with no weighting, uniform noise addition (0.01 and 0.02), and noise addition under a weighting scheme. Both adding obfuscating noise and down-weighting data points with high privacy risk push the ROC curve toward the random baseline. The combined effect of noise and weights significantly reduces the attacker's advantage, approaching random guessing as noise increases.}
\label{fig:comparison}
\end{figure*}

From the Figure \ref{fig:comparison} we observe that both adding obfuscating noise and down-weighting data points with high privacy risk have the effect of pushing down the ROC curve towards the random baseline.
Both the green solid line representing noise at 0.01 standard deviations and the weighted alternative (no noise) dashed blue line are well below the blue solid baseline. In the figure, the weighted results are from a scenario with parameters set to $\alpha=2$ and $\beta=2$, which were selected after noting that they give low privacy risk while maintaining acceptable accuracy (we will return to the analysis of the privacy-utility trade-off in the discussions that follow).

Furthermore, we see that the two effects compound -- the green dashed line representing the case where both noise and weights have been applied is well below each of the noise-only and weights-only alternatives.
As noise further increases to 0.03 standard deviations, the curves approach the random baseline, which we can interpret as that attacker advantage becomes no better than chance. 
\ifmycondition
  For details for each of the individual curves with confidence bands, we refer to Figure \ref{fig:confidence} in the Appendix, which shows that the difference in the random scenario is indeed not significant for the weighted and noise addition alternative (with noise addition at standard deviation 0.03).
\else
  For details for each of the individual curves with confidence bands, we refer to figures in the appendix of the accompanying technical report \cite{ForgottenByDesign2024}, which shows that the difference in the random scenario is indeed not significant for the weighted and noise addition alternative (with noise addition at standard deviation 0.03).
\fi

We make a similar observation from Table \ref{table:results_summary}, that is, privacy risk metrics AUC and tau drop off as we increase the magnitude of the obfuscating noise, and the effect is increased for the weighted alternative scenarios (here using same fixed weight parameters $\alpha$ and $\beta$ as for Figure \ref{fig:comparison}). 
As the noise standard deviation increases to the highest level of 0.05, we observe AUC close to 0.5 and tau close to 0.0, which indicates that an attacker has little advantage over random guessing. 

The tables also include test set accuracies, which are averaged over all shadow models and presented together with stdev in addition to mean (to permit error analysis). We observe that accuracy decreases with an increased level of obfuscation for both the uniform and the weighted scenarios. 

\begin{table}%[b]
\caption{Privacy Risk Metrics and Accuracy for Different Noise Levels and Weighting Schemes. The table presents the Area Under the Curve (AUC), tau at various False Positive Rates (FPR), and accuracy (mean and standard deviation) for different scenarios: baseline (no noise, no weights), uniform noise addition, and noise addition under a weighting scheme. The results show that both adding obfuscating noise and applying a weighting scheme significantly reduce privacy risks (lower AUC and tau values) while maintaining acceptable accuracy levels. As the noise level increases, the attacker's advantage diminishes, approaching random guessing. Numbers are based on experiments with at least 300 shadow models.}
\centering
\small
\begin{filecontents*}{Analysis/F/figures/results.csv}
,auc mean,tau@0.001FPR,tau@0.01FPR,tau@0.1FPR,acc mean,acc std,rho
uniform: 0.0,0.885,5.217,3.475,1.793,0.7,0.006,0.814
uniform: 0.01,0.756,3.745,2.521,1.294,0.697,0.004,0.915
uniform: 0.03,0.562,0.309,0.489,0.349,0.656,0.005,0.993
weighted: 0.0,0.879,4.525,3.345,1.802,0.684,0.009,0.917
weighted: 0.01,0.671,1.712,1.403,0.875,0.705,0.004,0.968
weighted: 0.03,0.543,0.159,0.292,0.226,0.643,0.005,0.991
\end{filecontents*}
\csvautobooktabular{Analysis/F/figures/results.csv}
\label{table:results_summary}
\end{table}

\begin{figure}%[H]
\centering
\includegraphics[width=\textwidth]{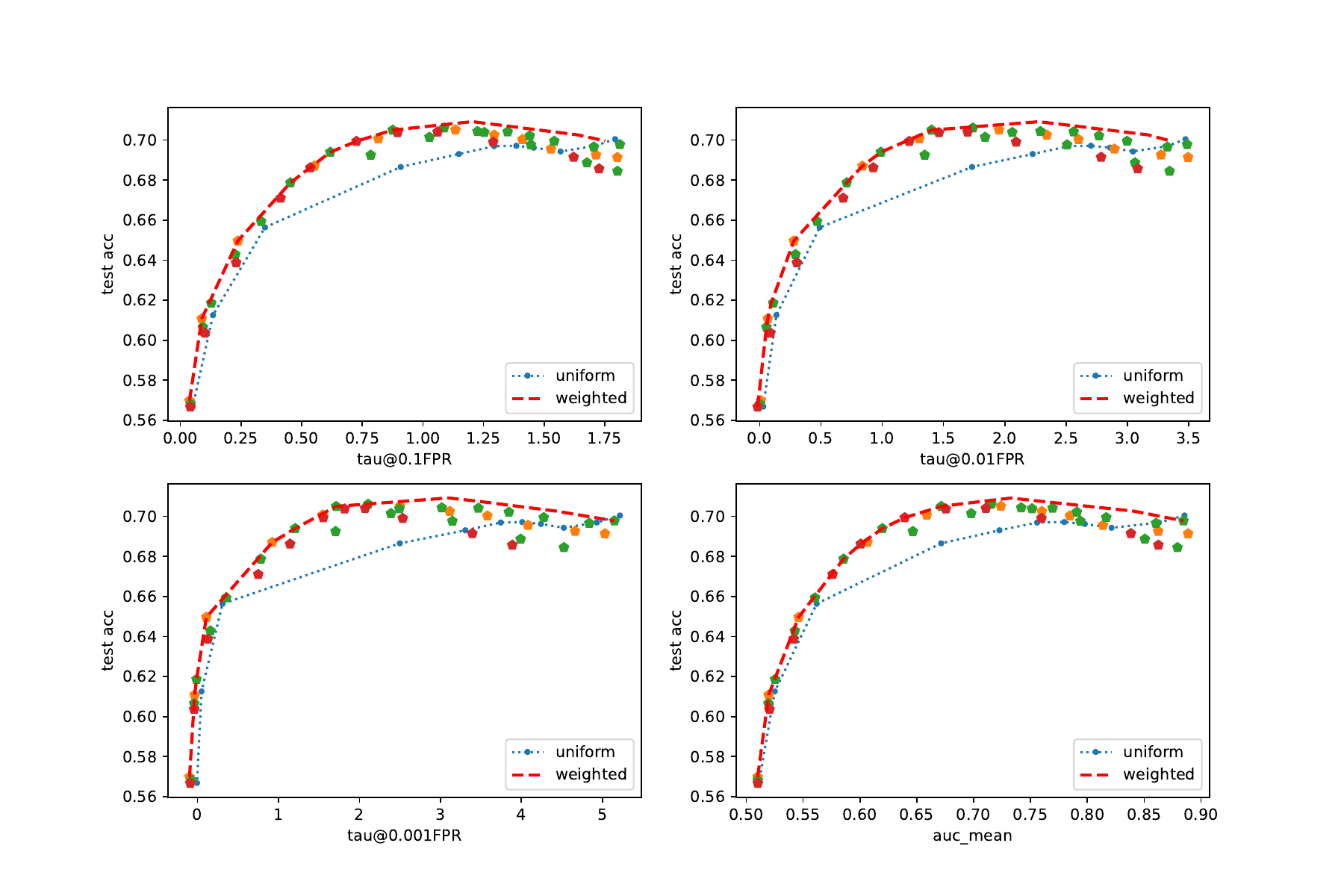}
\caption{Privacy-Utility Trade-Off for Different Noise Levels and Weighting Schemes. The scatter plots illustrate the relationship between privacy risk metrics, AUC and tau (log ratio of TPR over FPR), and model accuracy for different scenarios: baseline (no noise, no weights), uniform noise addition, and noise addition under a weighting scheme, where each dot is based on at least 300 shadow models. The curves show that as noise magnitude increases, privacy risk metrics decrease, initially with minimal (or even slightly positive) impact on accuracy. 
The red dashed line displays the convex hull for the weighted approach which demonstrates a more favorable accuracy for intermediate risk levels compared to the blue dotted line for uniform noise addition only, indicating an advantage in balancing privacy and utility.
}
\label{fig:trade-off}
\end{figure}

\paragraph{Privacy-Utility Trade-Off.}
To illustrate the relationship between privacy risk and accuracy, we can plot these in a scatter plot as in Figure \ref{fig:trade-off} -- this is the privacy-utility trade-off for our experiment, where we compare the resulting curves for the uniform case as the blue dotted line and the weighted case as the red dashed line (which is the convex hull for experiments taking different parameters $\alpha$ and $\beta$). Figure \ref{fig:trade-off} provides curves for tau at three FPR levels as well as for AUC (bottom right panel). 

The upper rightmost end of each curve corresponds to trials without noise addition, for which we observe both the highest values of the risk metrics and high accuracies. 
As the noise magnitude increases, the risk metrics decrease, at first, without much impact on accuracy as the curves are rather flat. As noise levels become material, we observe that accuracy drops materially.

In terms of the relationship between privacy risk metric and accuracy, the two curves separate, with the weighted alternatives having a more favorable accuracy for intermediate risk levels. For example, from Table \ref{table:results_summary} we observe that for the case setting noise standard deviation 0.02, the weighted case displays a tau=0.92 compared to the uniform case with tau=2.5 at 0.001 FPR, for the same accuracy. A similar observation can be made for the bottom right panel with AUC at 0.61 v.s.\ respectively. Overall, in each panel, we observe a range of values for the privacy risk metrics where the curves are clearly separated - this indicates that the weighted approach has an advantage over the version that only adds noise.
We even observe that accuracy can be higher compared to the non-obfuscated case for the weighted approach at some positive noise levels, for example, at $\sigma=0.01$ in Table \ref{table:results_summary}. We can speculate that it could be attributed to a regularization effect from the noise addition -- however, it is a small effect that is not statistically significant at 95\% in the table. 

Finally, towards the leftmost side, where the privacy risk metrics become smaller, the accuracy suffers significantly to a similar extent for both curves.
We note that while the privacy risk metrics for both approaches values that are representative of the random membership inference attack, accuracies remain well above those for a model that randomly assigns a label. For CIFAR-10, random label assignment would yield about 10\% accuracy compared to the strongly obfuscated cases in the sixties. 

\begin{figure*}%[H]
\centering
\includegraphics[width=\singlewidth]{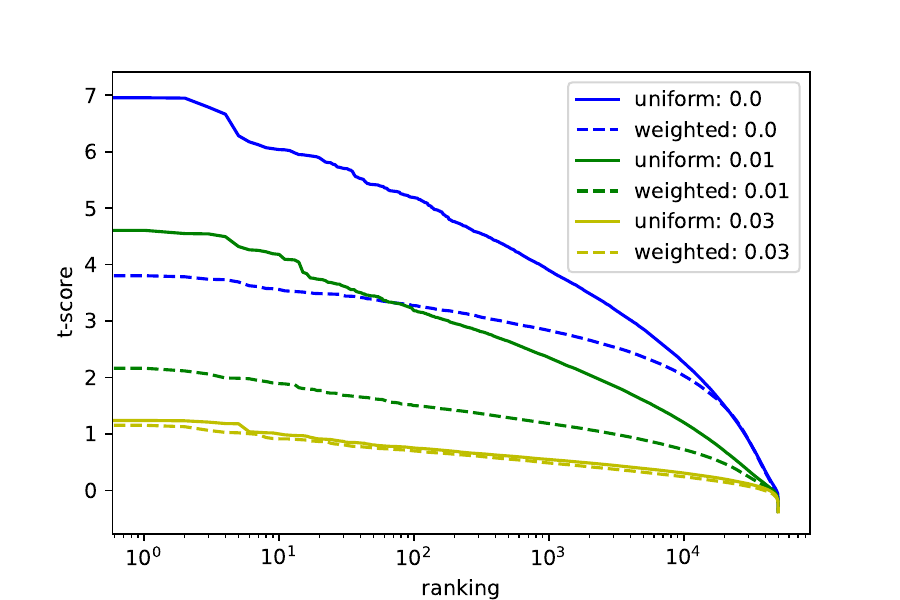}
\caption{Distribution of privacy vulnerability t-scores for Different Noise Levels and Weighting Schemes. The figure shows the distribution of t-scores for various scenarios: baseline (no noise, no weights), uniform noise addition, and noise addition under a weighting scheme. The results indicate that both noise addition and weighting shift the distribution of t-scores towards lower values, reducing the privacy risk. The combined effect of noise and weighting is particularly effective in lowering the highest privacy vulnerability t-scores.}
\label{fig:t_distance}
\end{figure*}

\paragraph{Privacy vulnerability t-scores.} 
Figure \ref{fig:t_distance} illustrates the distribution of the risk scores, calculated according to equation (\ref{eq:tscore}), for different choices of the noise parameter $\sigma$, where the blue solid line corresponds to the baseline model without any obfuscation. 
The risk scores are sorted in decreasing order and plotted with the ranking order on the x-axis. 
Dashed lines represent cases when the weighting has been applied and are always below the same colored solid lines that represent the cases without the weighting at the same noise standard deviation. 
The x-axis is on a log-scale to display more detail for the highest risk scores.

Taking the case when the weighting is not applied, we can compare the blue solid line ($\sigma=0.0$) with the green solid line ($\sigma=0.01$) in Figure \ref{fig:t_distance} to note that the noise addition shifts the distribution across the board, such both the highest risk scores are lowered (left side of the figure) as well as lower the scores for the bulk of the distribution (right side). We observe a similar shift increasing the noise level to $\sigma=0.03$ for the yellow solid distribution curve. 

Now, to the cases with and without the weighting, we can compare the blue solid line with the blue dashed line ($\sigma=0.0$ for both) to observe a shift towards lower privacy vulnerability t-scores. This effect is evident in the left side of the figure, where the highest score drops from almost 7 to below 4. Moving toward the right side of the figure, the effect is pronounced for the top 1000 to 10000 data points in the distribution but is diminished as we move into the bulk of the distribution, where the dashed and the solid curves eventually merge. 
A similar effect is observed in the presence of moderate noise ($\sigma=0.01$) for the green solid and dashed curves, while there is little difference at the higher noise level ($\sigma=0.03$, yellow curves) where privacy vulnerability t-scores are very low both with and without the weighting.
\ifmycondition
  Please refer to the appendix for a more detailed analysis of the privacy vulnerability t-score distributions (and comments to Figures \ref{fig:by_baseline_rank} and \ref{fig:in_vs_out}).
\else
  Please refer to the appendix of the accompanying technical report \cite{ForgottenByDesign2024} for a more detailed analysis of the privacy vulnerability t-score distributions.
\fi 

\section{Discussion}
\label{sec:discussions}

From Table \ref{table:results_summary}, which summarizes the results displayed in Figure \ref{fig:trade-off} for the privacy-utility trade-off, we see that the accuracy is not significantly different from the no-weights, no noise baseline ("uniform: 0.0"). Meanwhile, the drop in all the privacy risk metrics is material -- tau at 0.001 FPR is reduced from about 5.2 to about 1.7, which corresponds to a reduction of more than a magnitude in the expected attack rate success (measured as TPR over FPR). The effect of adding noise only ($\sigma=0.01$, no weights) is more modest, reducing the tau to 3.7 at the same FPR.

The analysis of the risk scores displayed in Figure \ref{fig:t_distance} is consistent with the observations we made for the privacy-utility trade-off of Figure \ref{fig:trade-off} -- adding noise and applying the weighting both lowers the privacy risk t-score as well as the observed risk metrics tau and auc. At lower noise levels, this only has a small effect on the utility (measured as accuracy). For a reference noise level of $\sigma=0.01$, where the TPR@0.01FPR dropped by an order of magnitude without significant impact on accuracy, we also observe in Figure \ref{fig:t_distance} a material downwards shift in the privacy vulnerability t-score distribution, with maximum values dropping from near 7 for the baseline ('uniform: 0.0') to slightly above 2 for the case with both noise addition and weighting applied ('weighted: 0.01'). 

\paragraph{Computational effort.} 
The weighting scheme uses t-scores to calculate the weights for each audit data point. This would require the model developer to train a sufficient number of shadow models to obtain t-scores before applying the weighting scheme to mitigate the privacy risk.
In comparison, noise addition without weighting, can be done without assessing the privacy risk of each data point and, therefore, doesn't require t-scores (or any other privacy risk metric). One could, therefore, conclude that the computational effort of applying the weighting is considerably higher than just using noise addition. 

A model auditor may require evidence that mitigations actually reduce the privacy risk of a model. One could, therefore, be expected to first test a baseline model by subjecting it to a membership inference attack -- this would require a first round of shadow models to be calculated to arrive at model auc and TPR at low FPR estimates. 
To assess the impact of the mitigations, one would then have to retrain a second round of shadow models with all mitigations applied.
In such a model audit scenario, at least two rounds of model testing would have to be carried out, each with the associated computational burden of training shadow models. Privacy vulnerability t-scores would be an output of the first round of training when the baseline is tested. These could then be used as a basis for the weighting mitigation for the target model, which is then tested in round two. Therefore, the computational effort would be similar for both the weighted and noise-only mitigation scenarios in such an audit scenario. 

\subsection{Relation to Previous Work}

\paragraph{Strong Adaptive Membership Inference Attacks. }
In their seminal work, Carlini et al. \cite{carlini2022membership} provided a comprehensive analysis of membership inference attacks, establishing foundational principles for evaluating and understanding these attacks. Our use of the LIRA membership inference attack builds on these principles, leveraging the robust methodology proposed by Carlini et al. to identify and protect vulnerable data points.
Recently, Zarifzadeh, Liu, and Shokri introduced RMIA \cite{rmia_zarifzadeh2023low} as a robust and computationally efficient method for performing membership inference attacks that maintain high accuracy and reliability even with limited reference models and data. It employs fine-grained modeling of the null hypothesis through many pairwise likelihood ratio tests.

\paragraph{Privacy Onion Effect. }
Carlini et al. \cite{carlini2022privacyonion} introduced the concept of the "Privacy Onion" to describe how removing the most vulnerable data points to privacy attacks exposes a new layer of data, which was previously considered safe, to similar risks.
This phenomenon underscores the complexity of ensuring data privacy in machine learning models. Our approach of using targeted obfuscation during training aligns with the need to address these layered vulnerabilities. By preventing sensitive data from being embedded in the model from the outset, our method that combines noise addition and down-weighting of vulnerable data points aims to mitigate the cascading risks highlighted by the Privacy Onion effect. 
\ifmycondition
  The effect on the privacy vulnerability t-score metric is analyzed further in the Appendix.
\else
  The effect on the privacy vulnerability t-score metric is analyzed further in the appendix of the accompanying technical report \cite{ForgottenByDesign2024}.
\fi 

\paragraph{Differential Privacy and Its Limitations. }
Differential Privacy (DP) has been widely studied as a defense against MIAs. However, as noted by Rahimian et al. \cite{rahimian2021differential}, DP often comes with a trade-off between privacy and model utility. Our approach of combining additive gradient noise with weighting schemes offers an alternative that aims to balance this trade-off more effectively. By tailoring the noise and weights to the specific vulnerability of data points, we strive to maintain model accuracy while enhancing privacy protection.

\paragraph{Empirical Defenses and Their Weaknesses. }
Recent evaluations by Aerni et al. \cite{aerni2024evaluations} have highlighted the weaknesses of empirical defenses that do not provide formal guarantees. 
Evaluations often use weak or non-adaptive membership inference attacks that do not fully exploit the vulnerabilities of the model. This leads to an underestimation of the privacy risks. For example, using simple threshold-based attacks rather than more sophisticated, adaptive attacks can result in misleading conclusions about the effectiveness of privacy defenses.
Our method only partly addresses these concerns; we use the preferred tail risk measure TPR at low FPR and give particular attention to vulnerable data points, however our work could be improved by considering the stronger RMIA attack and making a detailed comparison against state-of-the-art DP baselines.

\paragraph{Memorization and Privacy Risks. }
Choi et al. introduced the concept of memorization scores to measure how much a model's prediction improves when a data point is included in the training set. Our privacy vulnerability t-scores are conceptually similar, providing a metric to identify and mitigate the risks associated with memorization. By focusing on the most vulnerable data points, our approach aligns with the need to address the privacy risks highlighted by memorization studies.

\subsection{More on Limitations}

\paragraph{Generalizability to Other Datasets.} The current study is limited to the CIFAR-10 dataset. The effectiveness of the proposed methods may vary with different datasets, especially those with different characteristics or complexities. Future work should include experiments on a variety of datasets to validate the generalizability of the findings.
    
\paragraph{Scalability of Techniques.} The computational cost associated with training multiple shadow models and implementing instance-specific noise and weighting schemes may not scale well to larger datasets or more complex models. This could limit the practical applicability of the proposed methods in real-world scenarios where computational resources are constrained.
    
\paragraph{Impact on Model Performance.} While the proposed techniques aim to enhance privacy, they may also impact the overall performance of the model. The trade-off between privacy and utility needs to be carefully balanced, and further research is required to optimize this balance without incurring too large a computation cost.
    
\paragraph{Robustness to Advanced Attacks.} The study primarily focuses on membership inference attacks (and specifically only LIRA). However, there are other types of privacy attacks, such as model inversion and attribute inference attacks, that could also pose significant risks. The robustness of the proposed methods against these advanced attacks remains to be evaluated.
    
\paragraph{Assumptions in Privacy Metrics.} The privacy risk metrics used in this study, such as log-likelihood ratios and scaled logits, are based on certain assumptions about the distribution of data and model behavior. These assumptions may not hold in all cases, potentially affecting the accuracy of the privacy risk assessments.
    
\paragraph{Interpretation of RTBF under GDPR and Regulatory Acceptance.} The current definition of the right to be forgotten (RTBF) under GDPR is limited to erasing data, which our work contests. It is possible that changes in future legislation may not align with our suggested definitions, which address the current gap in RTBF for AI. Additionally, the regulatory acceptance of using membership inference attacks (MIA) for risk audits and the view on obfuscation techniques as risk mitigation strategies need further exploration. The alignment of these methods with evolving regulatory standards and their acceptance by regulatory bodies remains uncertain.
    
\paragraph{User Acceptance in Industry and Public Sector.} The adoption rate of privacy-preserving techniques also depends on how AI developers in industry and the public sector perceive the risk mitigation methods and the related testing methodology. The study does not explore the acceptance of these techniques as cost-efficient solutions to meet regulatory requirements related to model risk audits under GDPR and the AI Act. Future work could include user studies to assess the perception and willingness of AI developers to adopt these methods.

\section{Conclusions}
\label{sec:conclusions}
In this paper, we examined the concept of "Forgotten by Design" as a proactive strategy for privacy preservation in artificial intelligence (AI) systems. By incorporating obfuscation techniques during the model training process, we demonstrated the feasibility of preventing sensitive data from being embedded in AI models, thereby addressing the challenges associated with the "right to be forgotten" in the context of AI. Our experiments utilizing the LIRA membership inference attack on a ResNet-18 model trained on the CIFAR-10 dataset revealed that both noise addition and weighting schemes significantly mitigate privacy risks while maintaining acceptable model utility. The combined effect of these techniques approximates the performance of random guessing, effectively reducing the attacker's advantage.

The findings underscore the critical importance of integrating privacy-preserving mechanisms during the training phase of AI models. The proposed method aligns with human cognitive processes of motivated forgetting and demonstrates its potential as a robust framework for protecting sensitive information and ensuring compliance with privacy regulations. This research contributes to the advancement of privacy-preserving AI systems capable of adapting to evolving privacy standards and regulatory requirements.

\subsection{Future Work}

\paragraph{Adapting Methods for Intersectional Biases.} Conduct a pre-study to adapt the developed methods for mitigating intersectional biases in AI models. This will involve identifying and addressing potential biases that may arise from the interaction of multiple sensitive attributes.

\paragraph{Expand Dataset Variety.} Conduct experiments on a broader range of datasets with varying characteristics and complexities to validate the generalizability of the proposed methods.
    
\paragraph{Expanding Use Cases.} Select a use case of current interest, particularly one discussed by regulators such as the Swedish Authority for Privacy Protection (IMY)\footnote{https://www.imy.se/}, to demonstrate practical applicability and align with regulatory requirements.

\paragraph{Align with Regulatory Standards.} Engage with regulatory bodies to discuss the interpretation of RTBF under GDPR and the acceptance of membership inference attacks and obfuscation techniques as risk mitigation strategies, ensuring alignment with evolving standards.

\paragraph{Collaborative Projects.} Collaborate with industry and public sector partners to validate and refine our methods through systematic testing in various attack scenarios, including black-box, white-box, federated learning, and synthetic data. Additionally, assess the usability of the proposed methods for real-world tasks to ensure practical implementation.
    
\paragraph{Open-Source Development.} Develop and release an open-source tool based on our methods to provide scalable and relevant solutions for assessing and mitigating data leakage risks, for example, by integration into the LeakPro platform \cite{LeakPro2025}. This will support the broader community in designing privacy-preserving AI systems and preparing for future regulatory reporting requirements.

%% file: appendix.tex
\section*{Appendix}

This appendix presents a more detailed analysis of the t-scores as well as some full-page sized Tables and Figures that are referred from the main text. 

\subsection*{Additional Analysis of the t-scores}

\begin{figure*}[b]
\centering
\includegraphics[width=0.6\textwidth]{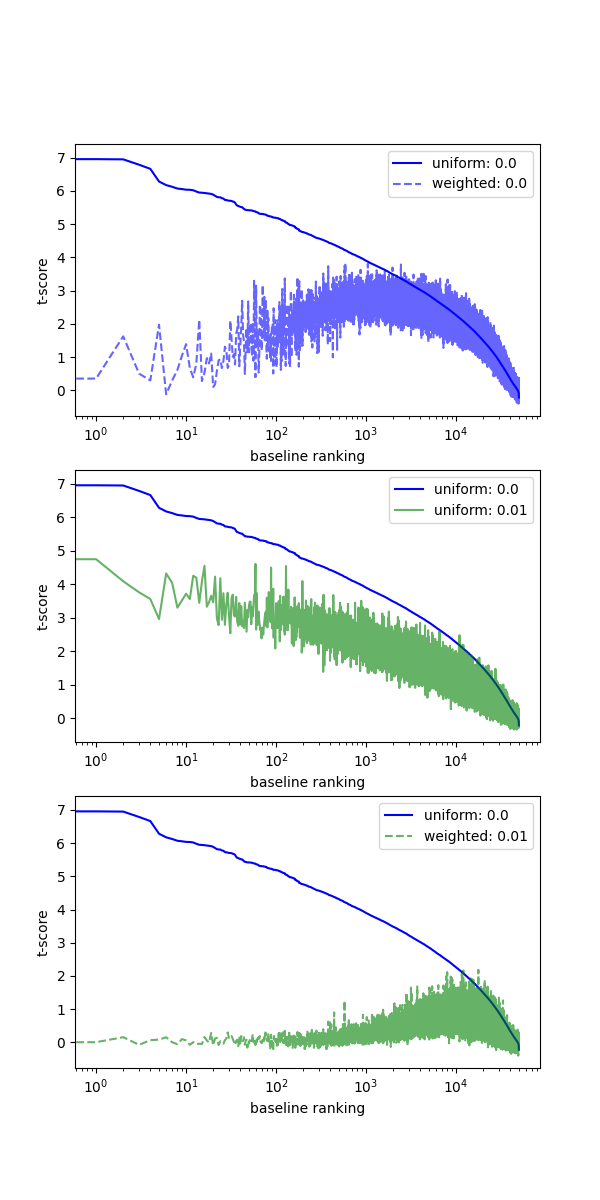}
\caption{Distribution of privacy vulnerability t-scores sorted by the original ranking from the baseline model. The figure shows the t-scores for various scenarios: baseline (blue solid line, $\sigma = 0.0$), noise addition only (green solid line, $\sigma = 0.01$), weighting only (blue dashed line, $\sigma = 0.0$), and both noise addition and weighting (green dashed line, $\sigma = 0.01$). The application of noise and weighting shifts the distribution towards lower t-scores, reducing privacy risk. This effect is particularly noticeable for the data points that have the highest original risk scores (i.e., without noise and weighting).}
\label{fig:by_baseline_rank}
\end{figure*}

Figure \ref{fig:by_baseline_rank} shows a different and more detailed view of how the noise addition and weighting scheme affect the privacy vulnerability t-score. It plots the distribution of privacy vulnerability t-scores for different noise levels and weighting schemes, sorted according to the ranking of the baseline model ("uniform: 0.0"), such that we can observe the changes in t-score for a point from the baseline value. 
Recall that the privacy vulnerability t-score, as defined in equation (\ref{eq:tscore}), is used to proxy the vulnerability of data points to membership inference attacks. It is calculated as the normalized difference between the mean of the IN and OUT distributions of scaled logits for each data point. 

The blue solid lines in each panel represent the baseline model without noise addition ($\sigma = 0.0$) and with no weighting applied, while the other lines represent specific model parameter alternatives. The top panel compares the noise-addition-only case in the green solid line ($\sigma = 0.01$, no weights). The mid panel, on the other hand, has no noise addition but applies the weighting scheme ($\sigma = 0.0$) in the dashed blue line. Finally, the lower panel displays the case when both noise addition ($\sigma = 0.01$) and weighting are applied in the green dashed curve.

In comparison to Figure \ref{fig:comparison}, which just sorted the t-scores within each scenario, which by construction gives naturally smooth curves, the plots now are considerably rougher as they incorporate the noise in the t-score estimation for each data point. There are, however, some clear patterns that emerge. 

From the figure, we observe that the application of noise shifts the distribution of t-scores towards lower values, reducing the privacy risk. This effect is evident across the entire distribution, with the highest risk scores being significantly lowered. For instance, the blue solid line ($\sigma = 0.0$) shows higher t-scores compared to the green solid line ($\sigma = 0.01$), indicating that noise addition effectively reduces the privacy risk.
The impact of the weighting scheme is also apparent. Comparing the blue solid line ($\sigma = 0.0$) with the blue dashed line ($\sigma = 0.0$ with weights), we see a shift towards lower t-scores, particularly for the highest risk scores. The combination of noise and weighting further reduces the privacy risk, as indicated by the lower t-scores in the dashed lines compared to their solid counterparts.

\subsection*{Further Dissection of the t-scores}

\begin{figure*}[b]
\centering
\includegraphics[width=\textwidth]{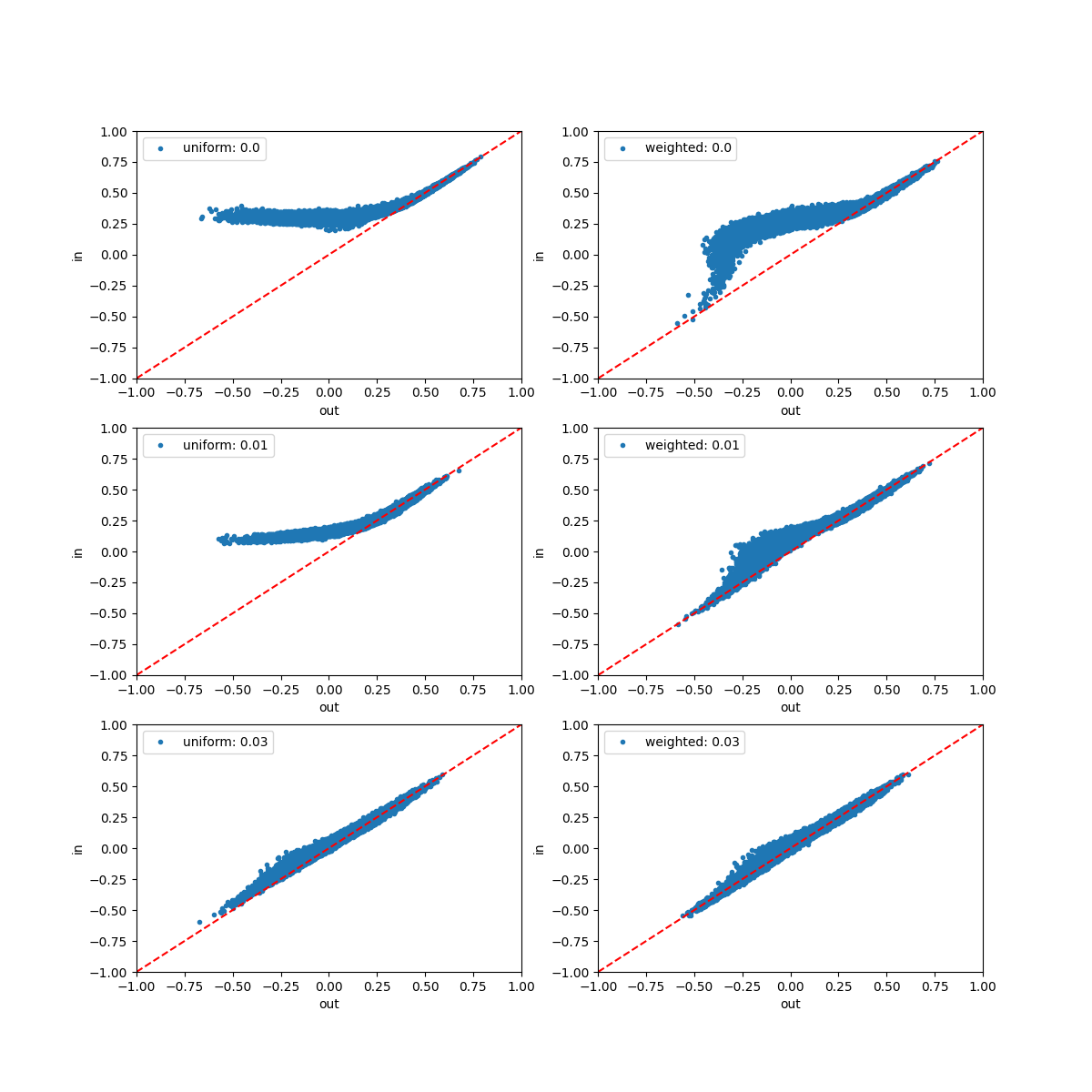}
\caption{Scatter plots of Normalized Means for IN and OUT Distributions at Different Noise Levels. The left panels show the scatter plots for three different levels of obfuscating noise (0.0, 0.01, and 0.03) without applying the weighting scheme. The right panels show the same noise levels with weights applied. The red dashed line represents the ideal case with no generalization gap. The application of noise and weighting schemes reduces the generalization gap, pushing the points toward the ideal model line and indicating reduced over-fitting and diminished opportunities for successful membership inference attacks.}
\label{fig:in_vs_out}
\end{figure*}

The t-score is related to the average generalization gap for an image in the audit dataset, i.e., the difference in performance for models where a specific image was in the train set or test set.  
We dissect the t-score by considering its two constituent terms -- in other words, we split it by the minus sign such that we have
\begin{equation}
t^{\IN}_i = \tanh{\left(\frac{m^{\IN}_j}{c_j}\right)}, \quad
t^{\OUT}_i = \tanh{\left(\frac{m^{\OUT}_j}{c_j}\right)}, \quad 
c_j = 2 \pi \sqrt{V^{\mathrm{IN}}_j+V^{\mathrm{OUT}}_j}
\label{eq:t-split}
\end{equation}
where the normalized means are first scaled by dividing by $2\pi$ and then passed through the hyperbolic tangent function. This is done to focus values around zero for visualizations like in Figure \ref{fig:in_vs_out} -- for comparison, these scatter plots include a red dashed line, which corresponds to the ideal case where there is no generalization gap and the model performs equally well (or bad) on both train and test data. 

The left panels show the scatter plots for three different levels of the obfuscating noise (0.0, 0.01, and 0.03) without applying the weighting scheme. Correspondingly, the plots on the right are for the same noise levels but with weights applied.

Zooming in on the baseline ("uniform: 0.0") we see that for points when the OUT models perform well, the IN model performance tends to be good -- here, the generalization gap is small and the scatter points align well with the dashed red diagonal line for the ideal model (upper right corner of the panel). As we slide down towards the lower left corner, we see signs of over-fitting, as IN models outperform OUT models, the curve a distinct banana-shaped upward bend. The application of noise pushes this bend down towards the ideal model line for the middle and lower (left) panels. This would also reduce the signal available through the t-score and diminish the opportunities for a successful LIRA attack. 

We can see the effect of the weighting scheme (without noise) in the upper right panel, which inhibits the over-fitting for the down-weighted points, pushing them towards the diagonal ideal model line. Adding noise on top of the weighting now further mitigates membership inference by reducing the signal for the bulk of points. We interpret this as an indication that both effects contribute to reduced memorization.  

\clearpage

\clearpage

\begin{figure*}
\centering
\includegraphics[width=\textwidth]{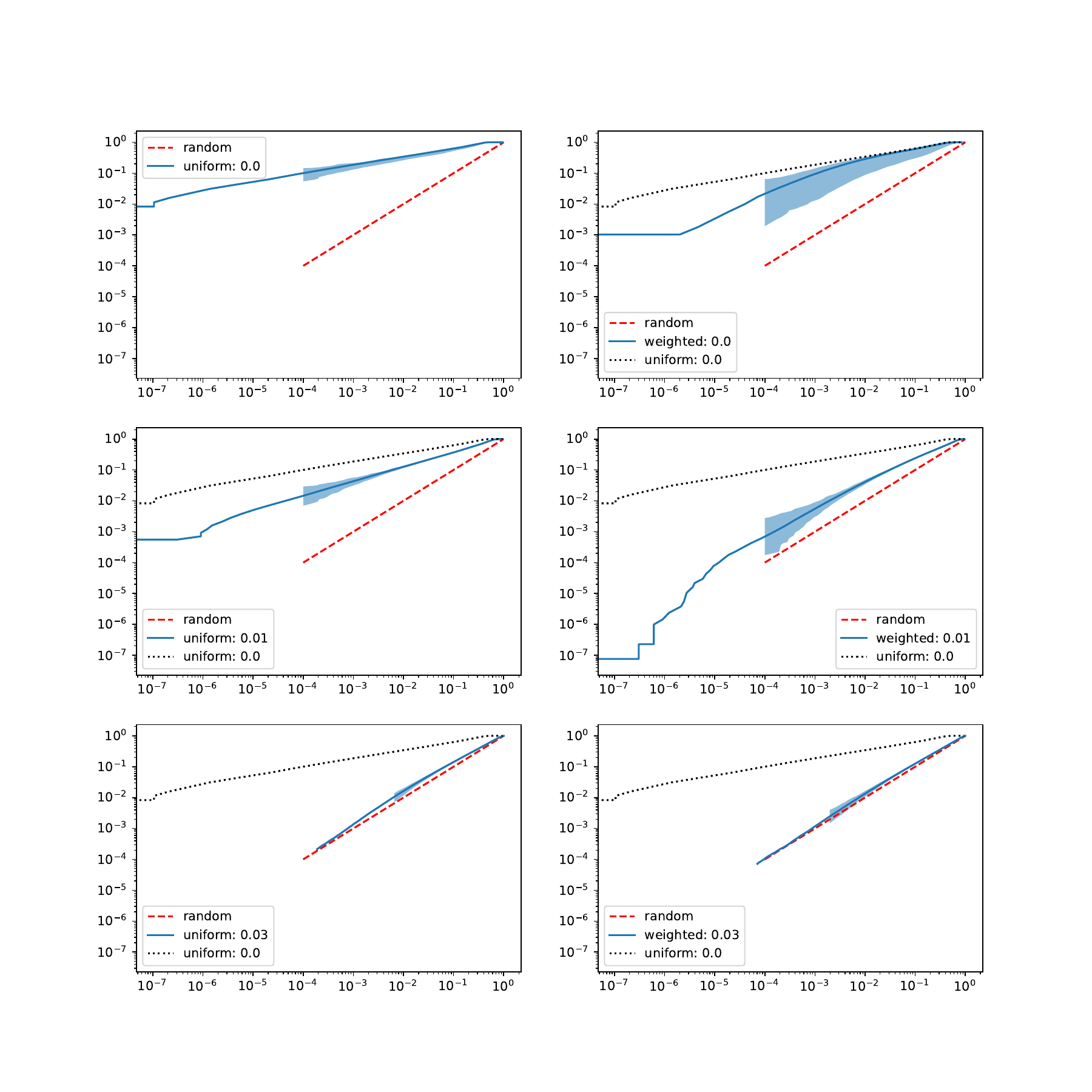}
\caption{ROC Curves with Confidence Bands for Different Noise Levels and Weighting Schemes. The figure shows that the difference to the random scenario is not significant for the weighted and noise addition alternatives with a noise standard deviation of 0.03.}
\label{fig:confidence}
\end{figure*}

\clearpage

\begin{table}%[b]
\caption{Privacy Risk Metrics and Accuracy for Different Noise Levels and Weighting Schemes. The table presents the Area Under the Curve (AUC), tau at various False Positive Rates (FPR), and accuracy (mean and standard deviation) for different scenarios: baseline (no noise, no weights), uniform noise addition, and noise addition under a weighting scheme.}
\centering
\small
\begin{filecontents*}{Analysis/F/figures/appendix.csv}
,auc mean,tau@0.001FPR,tau@0.01FPR,tau@0.1FPR,acc mean,acc std
uniform: 0.0,0.885,5.217,3.475,1.793,70.043,0.647
uniform: 0.001,0.863,4.932,3.322,1.703,69.693,0.865
uniform: 0.003,0.821,4.523,3.046,1.568,69.43,0.558
uniform: 0.005,0.798,4.241,2.859,1.457,69.622,0.459
uniform: 0.007,0.78,4.01,2.704,1.385,69.716,0.478
uniform: 0.01,0.756,3.745,2.521,1.294,69.693,0.443
uniform: 0.015,0.723,3.308,2.228,1.148,69.304,0.475
uniform: 0.02,0.671,2.498,1.734,0.909,68.655,0.467
uniform: 0.03,0.562,0.309,0.489,0.349,65.636,0.547
uniform: 0.04,0.525,0.049,0.136,0.134,61.255,0.633
uniform: 0.05,0.511,-0.006,0.029,0.051,56.682,0.826
weighted: 0.0,0.866,4.778,3.352,1.752,69.959,0.785
weighted: 0.001,0.841,4.433,3.168,1.637,70.262,0.694
weighted: 0.003,0.785,3.796,2.74,1.431,70.516,0.487
weighted: 0.005,0.755,3.432,2.481,1.301,70.773,0.431
weighted: 0.007,0.733,3.114,2.283,1.208,70.914,0.394
weighted: 0.007,0.732,3.077,2.27,1.198,70.914,0.39
weighted: 0.01,0.707,2.764,2.052,1.087,70.747,0.407
weighted: 0.015,0.684,2.328,1.763,0.97,70.201,0.397
weighted: 0.02,0.641,1.724,1.36,0.767,69.249,0.401
weighted: 0.03,0.56,0.354,0.468,0.333,65.933,0.486
weighted: 0.04,0.526,0.031,0.14,0.14,61.882,0.571
weighted: 0.05,0.51,-0.099,0.009,0.038,56.984,0.766
\end{filecontents*}
\csvautobooktabular{Analysis/F/figures/appendix.csv}
\label{table:results_appendix}
\end{table}